\documentclass[11pt,letterpaper]{article}
\usepackage{CJKutf8}

\usepackage{naaclhlt2013}
\usepackage{amsthm,amssymb,amsfonts}
\usepackage{mathtools}
\usepackage{latexsym}
\newcommand{\argmin}{\operatornamewithlimits{argmin}}

\newcommand{\PR}{\mathbf{Pr}}
\renewcommand{\P}{\mathbf{P}}
\makeatletter
\def\blfootnote{\xdef\@thefnmark{}\@footnotetext}
\makeatother

\title{Phrase Based Language Model for Statistical Machine Translation}

\author{Jia Xu and Geliang Chen*
\\
	    IIIS, Tsinghua University\\
	    {\tt xu@tsinghua.edu.cn, cglcgl200208@126.com}
\\— Working Paper —\\\today
}

\begin{document}
\begin{titlepage}

\maketitle

\blfootnote{* This version of the paper was submitted for review to EMNLP 2013. The title, the idea and the content of this paper was presented by the first  author in the machine translation group meeting at the MSRA-NLC lab (Microsoft Research Asia, Natural Language Computing) on July 16, 2013.}
\begin{abstract}

We consider phrase based Language Models (LM), which generalize the commonly used word level models. Similar concept on phrase based LMs appears in speech recognition, which is rather specialized and thus less suitable for machine translation (MT). In contrast to the dependency LM, we first introduce the exhaustive phrase-based LMs tailored for MT use. Preliminary experimental results show that our approach outperform word based LMs with the respect to perplexity and translation quality.


\end{abstract}

\section{Introduction}


Statistical language models estimating the distribution of
various natural language phenomena are crucial for many applications. In machine translation, it measures the  fluency and well-formness of a translation, and therefore is important for the translation quality, see ~\cite{Och:02} and ~\cite{Koehn:03} etc. 

Common applications of LMs include estimating the distribution based on N-gram coverage of words, to predict word and word orders, as in~\cite{Stolcke:02} and~\cite{crflm:01}. The independence assumption for each word is one of the simplifying method widely adopted. However, it does not hold in textual data, and underlying content structures need to be investigated as discussed in~\cite{Gao:04}.

\begin{CJK*}{UTF8}{gbsn}
\begin{table*}[htb]\label{fig-example}
  \begin{center}\scalebox{.9}{
  \begin{tabular}{cccccccc}\hline
     Words & John & played &basketball&the &day &before& yesterday\\
    $w_1^I$&$w_1$& $w_2$& $w_3$ & $w_4$&$w_5$&$w_6$& $w_7$\\
    Segmentations & \multicolumn{1}{c|}{John}&played & \multicolumn{1}{c|}{basketball}&the &day &before &yesterday   \\
     $k_1^J$       & \multicolumn{1}{c}{ $k_1=1$}&\multicolumn{2}{c}{ $k_2=3$}& \multicolumn{4}{c}{$k_3=7$}\\
$p_1^J$ & $p_1=w_1$&\multicolumn{2}{c}{$p_2=w_2 w_3$}&\multicolumn{4}{c}{$p_3=w_4 w_5$$w_6 w_7$}\\
    Re-ordered    &   John & \multicolumn{2}{|c|}{the day before yesterday }& \multicolumn{4}{c}{play  basketball} \\ 
Translation    &   约翰 & \multicolumn{2}{c}{昨天} & \multicolumn{4}{c}{打  篮球} \\\hline
  \end{tabular}}
\caption{\label{fig-example} Phrase segmentation example. }
  \end{center} 
\end{table*}
\end{CJK*}

We model the prediction of phrase and phrase orders. 
By considering all word sequences as phrases, the dependency inside a phrase is preserved, and the phrase level structure of a sentence can be learned from observations. This can be considered as an n-gram model on the n-gram of words, therefore word based LM is a special case of phrase based LM if only single-word phrases are considered. Intuitively our approach has the following advantages:

1) \emph{Long distance dependency}: 
The phrase based LM can capture the long distance relationship easily.  To capture the sentence level dependency, e.g. between the first and last word of the sentence in Table~\ref{fig-example}, we need a 7-gram word based LM, but only a 3-gram phrase based LM, if we take ``played the basketball'' and ``the day before yesterday'' as phrases.

2) \emph{Consistent translation unit with phrase based MT}: 
Some words may acquire meaning only in context, such as ``day", or ``the" in ``the day before yesterday" in Table~\ref{fig-example}. Considering the frequent phrases as single units will reduce the entropy of the language model. More importantly, current MT is performed on phrases, which is taken as the translation unit. The translation task is to predict the next phrase, which corresponds to the phrased based LM.

3) \emph{Fewer independence assumptions in statistical models}:
The sentence probability is computed as the product of the single word probabilities in the word based n-gram LM and the product of the phrase probabilities in the phrase based n-gram LM, given their histories. The less words/phrases in a sentence, the fewer mistakes the LM may contain due to less independence assumption on words/phrases. Once the phrase segmentation is fixed, the number of elements via phrase based LM is much less than that via the word based LM. Therefore, our approach is less likely to obtain errors due to assumptions. 

4) \emph{Phrase boundaries as additional information}:
We consider different segmentation of phrases in one sentence as a  hidden variable, which provides additional constraints to  align phrases in translation. Therefore, the constraint alignment in the blocks of words can provide more information than the word based LM.

\paragraph{Comparison to Previous Work}


In the dependency or structured LM, phrases corresponding to the grammars are considered, and dependencies are extracted, such as in ~\cite{Gao:04} and in ~\cite {Shen:08}. 
However, in the phrase based SMT, even phrases violating the grammar structure may help as a translation unit. For instance, the partial phrase ``the day before" may appear both in ``the day before yesterday" and ``the day before Spring". 
Most importantly, the phrase candidates in our phrase based LM are same as that in the phrase based translation, therefore are more consistent in the whole translation process, as mentioned in item 2 in Section 1.


Some researchers have proposed their phrase based LM for speech recognition. In ~\cite{Kuo:99} and~\cite{Tang:02}, new phrases are added to the lexicon with different measure function.  In~\cite{Heeman:97},  a different LM was proposed which derived the phrase probabilities from a language model built at the lexical level. 
Nonetheless, these methods do not consider the dependency between phrases and the re-ordering problem, and therefore are not suitable for the MT application.


\section{Phrase Based LM}


We are given a sentence as a sequence of words $w_1^I = w_1 w_2 \cdots w_i \cdots w_I (i\in 1,2,\cdots,I)$, where $I$ is the sentence length.

In the word based LM~\cite{Stolcke:02}, the probability of a sentence $\PR(w_1^I)$ \footnote{The notational convention will be as follows: we use the symbol $\PR$ to denote general probability distributions with (almost) no specific assumptions. In contrast, for model-based probability distributions, we use the generic symbol $\P(·)$.}is defined as the product of the probabilities of each word given its previous $n-1$ words:
\vspace{-.6cm} 
\begin{eqnarray}
\P(w_1^I)=\prod_{i=1}^I \P (w_i|w_{i-n+1}^{i-1})\label{eq-wordlm}
\end{eqnarray}
\vspace{-.4cm} 

The positions of phrase boundaries on a word sequence $w_1^I$ is indicated by $k_0\equiv 0$ and $K=k_1^J={k_1 k_2 \cdots k_j \cdots k_J }(j\in {1,2,\cdots,J})$, where $k_j \in \{1,2,\cdots,I\}, k_{j-1}<k_j, k_J \equiv I $, and $J$ is the number of phrases in the sentence. We use $k_j$ to indicate that the $j$-th phrase segmentation   is placed after the word $w_{k_j}$ and in front of word $w_{k_j+1}$, where $1\leq  j\leq J$. $k_0$ is a boundary on the left side of the first word $w_1$, which is defined as $0$, and $k_J$ is always placed after the last word $w_{I}$ and therefore equals $I$. 

An example is illustrated in Table~\ref{fig-example}. The English sentence ($w_1^I$) contains seven words ($I=7$), where $w_1$ denotes ``John", etc. The first phrase segmentation boundary is placed after the first word, and the second boundary is after the third word ($k=3$) and so on. The phrase sequence $p_1^J$ in this sentence have a different order than that in its translation, on the phrase level. Hence, the phrase based LM advances the word based LM in learning the phrase re-ordering.

\paragraph{(1) Model description}

Given a sequence of words $w_1^I$ and its phrase segmentation boundaries $k_1^J$, a sentence can also be represented in the form of a sequence of phrases $p_1^J=p_1 p_2 \cdots p_j \cdots p_J $$(j\in {1,2,\cdots,J})$, and each individual phrase $p_j$ is defined as 
\begin{eqnarray}
p_j=w_{k_{j-1}+1}\cdots w_{k_j} = w_{k_{j-1}+1}^{k_j} \nonumber
\end{eqnarray}
In phrase based LM, we consider the phrase segmentation $k_1^J$ as hidden variable and the Equation~\ref{eq-wordlm} can be extended as follows:
\begin{eqnarray}
\PR(w_1^I)&=&\sum_{K} \PR (w_1^I, K) \nonumber\\
&=& \sum_{{k_1^J},J} \PR(p_1^J | k_1^J)\cdot\PR(k_1^J)\label{eq-plm}
\end{eqnarray}

\paragraph{(2) Sentence probability}

For the segmentation prior probability, we assume a uniform distribution for simplicity, i.e. $\P(k_1^J )= 1/|K|$,  where the number of different $K$, i.e. $|K|=2^I$ if not considering the maximum phrase or phrase n-gram length; To compute the $\PR(w_1^m)$, we consider either two approaches:

\begin{itemize}
\item  \textsc{Sum Model} (Baum-Welch)

We consider all $2^I$ segmentation candidates. Equation~\ref{eq-plm} is defined as 
\begin{eqnarray}
\PR_{sum}(w_1^I)\approx \sum_{k_1^J,J} \prod_{j=1}^{J} \P(p_j  |  p_{j-n+1}^{j-1})\cdot \P(k_1^J)\nonumber,
\end{eqnarray}

\item \textsc{Max Model} (Viterbi)

The sentence probability formula of the second model is defined as
\begin{eqnarray}
\P_{max}(w_1^I)\approx \max_{k_1^J,J} \prod_{j=1}^{J} \P(p_j  |  p_{j-n+1}^{j-1})\cdot \P(k_1^J)\nonumber.
\end{eqnarray}
In practice we select the segmentation that maximizes the perplexity of the sentence instead of the probability to consider the length normalization.
\end{itemize} 
\paragraph{(3) Perplexity}
Sentence perplexity and text perplexity in the \textsc{sum model} use the same definition as that in the word based LM.
Sentence perplexity in the max model is defined as
$$PPL(w_1^I)= \argmin_{k_1^J,J} [\P(w_1^I , k_1^{J})]^{-1/J}\nonumber$$.
\vspace{-.5cm}

\paragraph{(4) Parameter estimation}

We apply maximum likelihood to estimate probabilities in both \textsc{sum model} and \textsc{max model} :
\begin{eqnarray}
\P(p_i |  p_{i-n+1}^{i-1})&=&  \frac{C(p_i)}{C(p_{i-n+1}^{i-1})}\label{eq-count1},
\end{eqnarray}
where $C(\cdot)$ is the frequency of a phrase.  The uni-gram phrase probability is $\P(p )= \frac{C(p)}{C}$, and $C$ is the frequency of all single phrases, in the training text. Since we generate exponential number of phrases to the sentence length, the number of parameters is huge. Therefore, we set the maximum n-gram length on the phrase level (note not the phrase length) as $N=3$ in experiments. 
\paragraph{(5) Smoothing}
For the unseen events, we perform Good-Turing smoothing as commonly done in word based LMs. Moreover, we interpolate between the phrase probability and the product of single word probabilities in a phrase using a convex optimization:
\begin{align*}
\P^* (p_j |  p_{j-n+1}^{j-1})=\qquad\qquad\qquad\qquad\qquad\qquad\;\;\;\;\\
\lambda \P(p_j |  p_{j-n+1}^{j-1})
+(1-\lambda )\frac{\prod_{i=1}^{j'}\P(w_i)} {\big(\sum_{w}\P(w)\big)^{j'}} \nonumber
\end{align*}
where phrase $p_j$ is made up of $j'$ words $w_1^{j'}$. The idea of this interpolation is to make the probability of a phrase consisting of of $j'$ words smooth with a $j'$-word unigram probability after normalization. In our experiments, we set $\lambda =0.4$ for convenience.

\paragraph{(6) Algorithm of calculating phrase n-gram counts}
The training task is to calculate n-gram counts on the phrase level in Equation~\ref{eq-count1}.

Given a training corpus $W_1^S$, where there are $S$ sentences $W_s$  ($s=1,2,\cdots,S$), our goal is to  to compute  $C(\cdot)$, for all phrase n-grams that the number of phrases is  no greater than $N$. Therefore, for each sentence $w_1^I$, we should find out every $n$-gram phrases that $0< n < N$.

We do Dynamic Programming to collect the phrase n-grams in one sentence $w_1^I$:
\begin{align*}
&Q(1,d;w_1^I)=\{p=w_b^d , \forall 1\leq b\leq d \leq I\}\\
&Q(n,d;w_1^I)=\\
& \cup_b Q(n-1, b-1;w_1^I)\oplus p=w_{b}^d,\; \forall n\leq b \leq d \leq I ,
\end{align*}
where $Q(\cdot)$ is the auxiliary function denoting the multiset of all phrase n-grams or unigram ending at position $d$ ($1<n\leq N$). $b$ denotes the starting word position of the last phrase in the multiset. The $\{\cdot\}$ is a multiset, and $\oplus$ means to append the element to each element in the multiset. $\cup_b$ denotes the union of multisets. After appending $p$,  we consider all $b$ that is no less than $n$ and no greater than $d$. 

The phrase counts $C(\cdot)$ is the sum of all phrase n-grams from all sentences $W_1^S$, with each sentence $W_s=w_1^I$, and $|\cdot|$ is the number of elements in a multiset:
\begin{eqnarray}
C(p_1^n)=\sum_{s=1}^S|p_1^n\in \cup_{d=n}^{|W_s|}Q(n,d;W_s)| \nonumber
\end{eqnarray}

\section{Experiments}

\begin{table}[t]
\begin{center}\scalebox{.9}{
\begin{tabular}{|l|c|c|c|} \hline
Data	&Sentences&	Words&	Vocabulary\\\hline
Training	&54887	&576778	&23350\\\hline
Dev2010	&202	&1887	&636\\\hline
Tst2010	&247	&2170	&617\\\hline
Tst2011	&334	&2916	&765\\\hline  
\end{tabular}}
\end{center}
\caption{\label{tab-data} Statistics of corpora with sentence length no greater than 15 in training and 10 in test. }
\end{table}

\begin{table}[t]
\begin{center}\scalebox{.9}{
\begin{tabular}{|l|c|c|c|c|c|}\hline
 n&Base&	\textsc{Sum}&	\textsc{Sum}+S.&	\textsc{Max}&	\textsc{Max}+S. \\\hline
1   & 676.1   & 85.5  & 112.5 &   625.7  &  1129.4\\\hline
2   &180.8   & 52.6  &  72.1   & 161.1  &  306.2\\\hline
3   & 162.3 &  52.5  &  72.2 &   140.4   & 266.5\\\hline
4    &162.5   & 52.6  &  72.3   & 141.1  &  267.6\\\hline
\end{tabular}}
\end{center}
\caption{\label{tab-ppl} Perplexities on Tst2011 calculated based on various n-gram LMs with $n=1,2,3,4$. }
\end{table}

\begin{table}[t]
\begin{center}\scalebox{.9}{
\begin{tabular}{|l|c|c|c|} \hline
Model	&Dev2010	&Tst2010	&Tst2011\\\hline
Base	&11.26	&13.10	&15.05\\\hline
Word	&11.92	&12.93	&14.76\\\hline
\textsc{Sum}	&11.86	&12.77	&14.80\\\hline
\textsc{Sum}+S.	&12.02	&12.54	&14.76\\\hline
Max	&11.61	&12.99	&15.34\\\hline
\textsc{Max}+S.	&11.56	&13.55	&15.27\\\hline
\end{tabular}}
\end{center}
\caption{\label{tab-bleu} Translation performance on N-best list using different LMs  in  BLEU[\%]. }
\end{table}

\begin{table}[t]
\begin{center}\scalebox{.9}{
\begin{tabular}{ll} \hline
Base: &but we need a success\\
   \textsc{Max}: &but we need a way to success .\\
   Ref: &we certainly need one to succeed .\\\hline
 Base: &there is a specific steps that\\
   \textsc{Max}: &there is a specific steps . \\
   Ref: &there is step-by-step instructions on this .\\\hline
\end{tabular}}
\end{center}
\caption{\label{tab-outputs} Examples of sentence outputs with baseline method and with the \textsc{max model}. }
\end{table}

This is an ongoing work, and we performed preliminary experiments 
on the IWSLT~\cite{iwslt:11} task, then evaluated the LM performance by measuring the LM perplexity and the MT translation performance.
Because of the computational requirement, we only employed sentences which contain no more than 15 words in the training corpus and no more than 10 words in the test corpora (Dev2010, on Tst2010 and on Tst2011), as shown in Table~\ref{tab-data}.  

We took word based LM in Equation~\ref{eq-wordlm} as the baseline method (Base). 
We calculated the perplexities of Tst2011 with different n-gram orders using both \textsc{sum model} and \textsc{max model}, with and without smoothing (S.) as in Section~2. 
Table~\ref{tab-ppl} shows that perplexities in our approaches are all lower than those in the baseline. 

For MT, we selected the single best translation output based on the LM perplexity of the 100-best translation candiates, using different LMs as 
 shown in Table~\ref{tab-bleu}. \textsc{Max~model} along with smoothing outperforms the baseline method under all three test sets with  the BLEU score~\cite{Papineni:02} increase of 0.3\% on Dev2010, 0.45\% on Tst2010, and 0.22\% on Tst2011, respectively.

Table~\ref{tab-outputs} shows two examples from the Tst2010, where we can see that our \textsc{max model} generates better selection results than the baseline method in these cases.

\section{Conclusion}
We showed the preliminary results that a phrase based LM can improve the performance of MT systems and the LM perplexity. We presented two phrase based models which consider phrases as the basic components of a sentence and perform exhaustive search. Our future work will focus on the efficiency for a larger data track as well as the improvements on the smoothing methods. 


\end{titlepage}

\end{document}